\documentclass[10pt,twocolumn,letterpaper]{article}

\usepackage{cvpr}
\usepackage{times}
\usepackage{epsfig}
\usepackage{graphicx}
\usepackage{amsmath}
\usepackage{amssymb}
\usepackage{caption}
\usepackage{float}

\graphicspath{ {Images/} }

\usepackage[pagebackref=true,breaklinks=true,letterpaper=true,colorlinks,bookmarks=false]{hyperref}

 \cvprfinalcopy 

\ifcvprfinal\fi
\begin{document}

\title{Robust Shape Registration using Fuzzy Correspondences}

\author{A Kolagunda, S Sorensen, P Saponaro, W Treible and C Kambhamettu\\
University of Delaware, Newark, DE 19711, USA\\
{\tt\small https://www.eecis.udel.edu/wiki/vims}
}

\maketitle

\begin{abstract}
    Shape registration is the process of aligning one 3D model to another. Most previous methods to align shapes with no known correspondences attempt to solve for both the transformation and correspondences iteratively. 
We present a shape registration approach that solves for the transformation using fuzzy correspondences to maximize the overlap between the given shape and the target shape. A coarse to fine approach with Levenberg-Marquardt method is used for optimization. Real and synthetic experiments show our approach is robust and outperforms other state of the art methods when point clouds are noisy, sparse, and have non-uniform density. Experiments show our method is more robust to initialization and can handle larger scale changes and rotation than other methods. We also show that the approach can be used for 2D-3D alignment via ray-point alignment.
\end{abstract}

\section{Introduction}
Registration is a common problem in many 3D tasks and has wide ranging applications to scene understanding, modeling, robotics, and many other research areas. The goal is to find the transformation that aligns two objects. In this paper we focus on model fitting where we align a given partial point set to a model represented as a point set or a mesh. While there are many approaches for registration, they often suffer greatly in cases where the source and target models vary greatly due to scale changes, noise, outliers and missing data.

\begin{figure}
\includegraphics[width=0.45\textwidth]{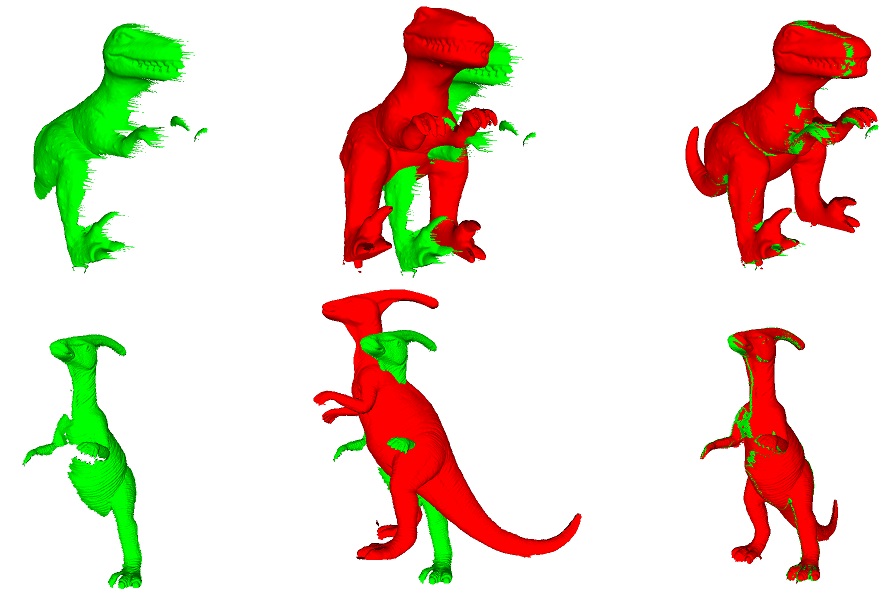}
\caption{Sample alignment results of objects from \cite{Mian2006,1677516}. The left side shows the target shape, the center shows the initialization, and the right shows our results.}
\label{fig:dino_alignment}
\end{figure}

We propose a novel method for registration which solves for the transformation that maximizes the proximity of a given shape to a target shape while also maximizing coverage on the target shape. Proximity is a measure of the distance between the two shapes. Coverage is the area of the target shape that corresponds to the given shape. This approach does not require correspondences, is robust to scale changes, large transformations, and works well on non-uniform points making it ideal for many real world applications. A sample of alignment results using our method is shown in Figure \ref{fig:dino_alignment}. We demonstrate our technique on a variety of 3D benchmarks including a typical  registration dataset, a dataset of low resolution, non-uniform and noisy LiDAR scans with corresponding CAD models, and a series of 3D to 2D alignment experiments. The technique outperforms traditional registration approaches on standard 3D alignment, and we demonstrate its wide ranging applicability by illustrating its use for 3D to 2D projective alignment to calibrate a camera system and LiDAR. 

The paper is organized as follows. Section \ref{sec:rel_works} describes related registration algorithms. Section \ref{sec:method} describes our energy function, the proximity and coverage metrics and fuzzy correspondences. Section \ref{sec:2d_3d} describes 2D to 3D alignment through point-to-ray registration. Section \ref{sec:exp} describes our experiments including a direct comparison to relevant methods on the Correspondence and Registration of Range Images for 3D Modeling dataset, the Ford Campus dataset, and 2D-3D alignment on our own dataset. Finally, we conclude the paper in Section \ref{sec:con}

\subsection{Related Works}
\label{sec:rel_works}
3D registration is well studied problem with many different approaches, we direct the reader to \cite{castellani20123d} for a more comprehensive review. ICP \cite{journals/ivc/ChenM92,121791} is the most widely used method for rigid registration of point sets. ICP is an iterative method that alternates between two steps (i) computing correspondences with fixed transformation and (ii) solving for the transformation with fixed correspondences. Closest points between the point sets are assigned as correspondences. ICP works well when the transformation between the point sets is small and the data has low levels of noise, few outliers, and uniform density of points. Many variants have been proposed to overcome limitations of ICP \cite{conf/icip/ZinsserSN03,citeulike:937464,conf/3dim/GelfandRIL03,conf/3dim/JostH03,low2004linear}.\cite{low2004linear} use a point to plane distance to find correspondences,\cite{conf/3dim/JostH03} use a hierarchical approach and propose a heuristic points search algorithm to speed up ICP, \cite{conf/3dim/GelfandRIL03} propose a method to sample geometrically stable points used to solve for transformations.

To increase robustness to outliers and noise, \cite{Rangarajan_softassign} proposed to soft assign correspondences based on Gaussian weights, \cite{Granger_EM-ICP} proposed a probabilistic approach EM-ICP in which they used matches weighted by normalized Gaussian weights. Though they replaced the discrete matching by a continuous function their methods still alternate between finding correspondences and solving for transformations. \cite{fitzgibbon2003robust} proposed LM-ICP which uses nonlinear optimization technique such as Levenberg-Marquardt method to solve for both correspondences and transformation in the same step. They replace discrete matching by a kernel function of distance transform. 

\cite{Breitenreicher2010}  propose a continuous representation of point sets using a mixture model and minimize the KL distance between the point sets using an improved Euler's algorithm for numerical integration. These methods, while being robust to noise and outliers, are still only well suited for point sets of similar scale. Myronenko and Song \cite{5432191} proposed Coherent Point Drift (CPD), a probabilistic approach for rigid and non-rigid registration. They represent the point set using GMM and force the centroids to move coherently for preserving topology. They also propose a close form solution to the EM approach to solve for the transformation. 

\section{Methods}
\label{sec:method}
Our approach to align shapes represented as point sets without explicit correspondences involves solving for the transformation $\theta$ of a given shape $D$ that maximizes the proximity of $\theta(D)$ to a model $S$ while maximizing coverage on $S$. In this paper,  we solve for rigid and similarity transformations. We hypothesize that a given shape $D$ is aligned to a model $S$ if all the points of $D$ are within a distance $\epsilon$ from $S$ (Proximity) and if all the points of $S$ are within a distance $\epsilon$ from $D$ (Coverage). When $D$ and $S$ are identical sets this amounts to $one-to-one$ mapping between them. But, in most real world scenarios, these sets are not identical. The two point sets differ in cardinality and point density along with the data being corrupted by noise, outliers, and missing data. In such scenarios we propose to maximize proximity and coverage by minimizing the following energy function:     

\begin{equation}
\label{eq:Energy}
E(D,S,\theta) = \alpha (1- P(\theta D,S)) + \beta (1- C(\theta D,S))
\end{equation}

where, $P$ and $C$ are measures of proximity and coverage respectively. 
$\alpha$ is the weighting factor in the range [0,1] and $\beta = 1-\alpha$.
\begin{gather}
P(D,S) = \frac{1}{|D|}\sum_{x_i \in D} \delta(x_i,S) \label{eq:P1}\\
C(D,S) = \frac{1}{|S|}\sum_{y_j \in S} \delta(y_i,D) \label{eq:C1}\\
\delta(x,Y) = \begin{cases}
      1, & \text{if}\ \exists y_j \in Y\ s.t.\ |x-y_j| < \epsilon  \\
      0, & \text{otherwise}
    \end{cases}
\end{gather}

The discrete functions $P$ and $C$ do not allow minimization of equation \eqref{eq:Energy} using standard numerical nonlinear optimization methods. Note that, given an $\epsilon$, by minimizing equation \eqref{eq:Energy} we are optimizing the mapping (correspondences) between the point sets rather than directly minimizing the distance between the point sets. $\delta$ does not measure the quality of mapping between the point sets. Inspired by \cite{Rangarajan_softassign,Granger_EM-ICP,fitzgibbon2003robust,Breitenreicher2010} which use weighted matching of points to make the correspondence matrix continuous, we replace the function $\delta$ by a summation of Gaussian functions. We define a fuzzy correspondences matrix $M$ between point sets $D$ and $S$ as

\begin{equation}
M_{i,j} = e^{\frac{-||x_{i} - y_{j}||^2}{2\sigma^2}}
\end{equation}
where $x_{i} \in D$ and $y_{j} \in S$. A matrix element $M_{i,j}$ gives a measure of quality of the mapping between $x_i \in D$ and $y_j \in S$. The $\sigma$ here is analogous to $\epsilon$. 
Using the fuzzy membership matrix $M$ we redefine Proximity $P$ and Coverage $C$ as
\begin{gather}
P(D,S) = \frac{1}{|D|}\sum_{i=1}^{|D|}\sum_{j=1}^{|S|}M_{i,j} \label{eq:P2} \\
C(D,S) = \frac{1}{|S|}\sum_{j=1}^{|S|}\sum_{i=1}^{|D|}M_{i,j}. \label{eq:C2}
\end{gather}

Since the cardinality of the two sets are generally not the same (M is not a square matrix), the inner summation in equations \eqref{eq:P2} and \eqref{eq:C2} may bias the energy equation \eqref{eq:Energy}  towards $P$ when $|D| \ll |S|$ and towards $C$ when $|S| \ll |D|$. Replacing the inner summation by an averaging operation will eliminate this bias, however it can lead to small values which can result in numerical instability during optimization. We circumvent these issues by introducing normalized fuzzy correspondences. We define normalized fuzzy correspondences matrices $M^P$ and $M^C$ to compute $P$ and $C$ respectively.

\begin{gather}
M^{P}_{i,j} = \frac{M_{i,j}}{\sum_{y_l \in S}e^{\frac{-||y_j-y_l||^2}{2\sigma^2}}} \\
M^{C}_{i,j} = \frac{M_{i,j}}{\sum_{x_l \in D}e^{\frac{-||x_i-x_l||^2}{2\sigma^2}}}
\end{gather}

This normalization however does not guarantee that the inner summation in equations \eqref{eq:P2} and \eqref{eq:C2} is always $<1$. We use a log-sigmoid function to enforce this constraint. We can now calculate $P$ and $C$ as
\begin{gather}
P(D,S) = \frac{1}{|D|}\sum_{i=1}^{|D|}\frac{1}{1+e^{-k\sum_{j=1}^{|S|}M^{P}_{i,j}}} \label{eq:P3} \\
C(D,S) = \frac{1}{|S|}\sum_{j=1}^{|S|}\frac{1}{1+e^{-k\sum_{i=1}^{|D|}M^{C}_{i,j}}} \label{eq:C3}
\end{gather}

\par
$k$ is a constant that is set so that we have well scaled gradients for optimization. We set $k=2$ in our experiments. The Gaussian scale $\sigma$ controls the capture range of a model. Points that are outside the capture range of a model do not contribute to the energy function. Therefore, large $\sigma$ makes the method robust to initialization, while small $\sigma$ makes it robust to outliers. We describe a multi-$\sigma$ approach for registration in the following section.
\subsection{Optimization}
\label{sec:optimization}
The energy function in equation \eqref{eq:Energy} is minimized using the Levenberg-Marquardt method. The LM method is used to solve least-squares fitting problems. The energy function \eqref{eq:Energy} can also be written as a sum of residuals
\begin{gather}
E^*(\theta) = \sum_{l}^{N}e_{l}^2(D,S,\theta) \label{eq:Ef}
\end{gather}
where $N = |D| + |S|$ and $e$ is the vector
\begin{gather*}
e(D,S,\theta) = \{\frac{\alpha}{|D|} (1- P(\theta x_i,S)) | x_i \in D\}\ 
\cup \\
\{\frac{\beta}{|S|} (1- C(\theta D,y_j)) | y_j \in S\} 
\end{gather*}
The error function \eqref{eq:Ef} is minimized to solve for the transformation $\theta$. The Jacobian required by the method is computed $J_{k,l} = \frac{\delta e_{k}}{\delta \theta_{l}}$, where $1\leq k \leq N$, $1\leq l \leq N_{\theta}$. $N_{\theta} = 7$ for rigid transformation and $N_{\theta} = 8$ for similarity transformation.
We use unit quaternions to represent rotation. To be robust to initialization we start the optimization at larger $\sigma$ values and iteratively start reducing the value of $\sigma$. We scale the point sets so that they fit within an unit cube and after every optimization step reduce $\sigma$ by a factor of $2$ (The initial and final values for $\sigma$ are set based on the application). Finally the point sets are scaled back to the original scale of the target shape. We also use a multi-resolution approach where we iteratively increase the cardinality of the two point sets. Initially when the $\sigma$ is large, a point in $D$ can be mapped to multiple points in $S$. As the $\sigma$ is reduced, the mapping moves towards a one-to-one mapping between the point sets with outliers having no mapping. This builds in an outlier rejection mechanism into the registration process while being robust to initialization.

\subsection{2D-3D alignment}
\label{sec:2d_3d}
Here we extend the point set registration scheme to achieve 2D to 3D alignment through point-to-ray registration. Given a set of 2D image points and a set 3D scene points that represent objects in the scene, we wish to solve for the transformation that aligns the projection of the 3D points with 2D image points. We formulate this problem as solving for the rigid transformation that aligns 3D points $D$ to rays $R$. From the 2D image points, using the camera projection parameters, we compute a set of normalized rays $R$ that originate from the center of projection and pass through the 2D image points on the image plane. Assuming that the center of projection is at the origin, We modify the fuzzy correspondence matrix by replacing the point-to-point distance by point-to-ray distance.

\begin{gather}
M_{i,j} = exp^{\frac{x_i \cdot r_j - x_{i}^{2}}{2\sigma^2}} \\
M^{P}_{i,j} = \frac{M_{i,j}}{\sum_{r_l \in R}exp^{\frac{r_l \cdot r_j - r_{l}^{2}}{2\sigma^2}}} \\
M^{C}_{i,j} = \frac{M_{i,j}}{\sum_{x_l \in D}exp^{\frac{x_l \cdot norm(x_i) - x_{i}^{2}}{2\sigma^2}}}
\end{gather}
Where, $r_j \in R$ and $x_i \in D$. Optimization of the alignment is performed as in section \ref{sec:optimization}. In the experiments section we show the application of this method to align points from LiDAR to image points.
\subsection{Voxelization}
\label{voxel}
The proposed approach requires that the two shapes $D$ and $S$ be represented as point sets. In cases when a shape is input as a mesh with low vertex density we perform voxelization to convert them to point sets. To voxelize a mesh, we use triangle-cube intersection to find voxels that intersect a face in the mesh. We initialize a voxel grid of desired resolution within the bounding box of the shape. The voxels which intersect any face of the mesh are set to $1$ and the ones that do not are set to $0$. This gives us a binary voxel representation of the surface of the shape. We then use morphological closing to remove small gaps. To mask out voxels that are not part of the outer surface of the shape, we set all the voxels that are not reachable from the edge of the volume to $0$. The centers of all the voxels that are set to $1$ form the point set representation of the shape.

\section{Experiments}
\label{sec:exp}
To validate our method we have conducted three sets of experiments on diverse data types. First we compare our method directly to common registration schemes on the Automatic Correspondence and Registration of Range Images for 3D Modeling dataset \cite{Mian2006,1677516}, which is a publicly available 3D dataset for object recognition, segmentation and registration. Secondly, to demonstrate the efficacy of our approach on noisy and sparse data with scale differences we use the Ford Campus Dataset \cite{Bao_CVPR2011_SSFM}, which features noisy LiDAR scans of a parking lot and align these to high quality  CAD models from 3D Warehouse.  We also present results of 2D to 3D registration in the form of LiDAR to camera alignment, to demonstrate the generality of our technique. The result is an automatic or semi-automatic approach to LiDAR to camera calibration which does not require a calibration object.

\subsection{Registration comparison}
The Automatic Correspondence and Registration of Range Images for 3D Modeling dataset \cite{Mian2006,1677516} features 4 different 3D models in various scenes that have been scanned with a non contact laser scanner. The dataset features high resolution $360^{\circ}$ scans of each object on its own as well as scans of scenes containing combinations of the objects in different arrangements and poses. The dataset has been developed for a number of tasks including segmentation and detection, but the cluttered scenes contain ground truth alignment parameters, allowing us to validate our approach by registering the high resolution target shape to the occluded or translated objects in scanned scenes.  

Since we are only concerned with registration and not segmentation or detection, we semi-automatically segment out  each shape in the scenes containing multiple objects. We start with an aligned shape and we rotate the source shape in steps of five degrees along each axis. We then align the centroid of both models, introducing a translation, as the target shape is incomplete. We compare our approach to a number of common registration schemes, including ICP  \cite{journals/ivc/ChenM92,121791} with point to plane distance, CPD \cite{5432191}, Efficient variants of ICP \cite{conf/icip/ZinsserSN03,citeulike:937464,conf/3dim/GelfandRIL03,conf/3dim/JostH03,low2004linear}, ICP Fast\cite{Mian2006,1677516}, Finite-ICP \cite{kroon} and compare the resulting alignment parameters against the ground truth by measuring mean vertex to vertex distance.

\begin{figure*}
\centering
\begin{minipage}[c]{0.33\textwidth}
\centering
\includegraphics[width=1\textwidth]{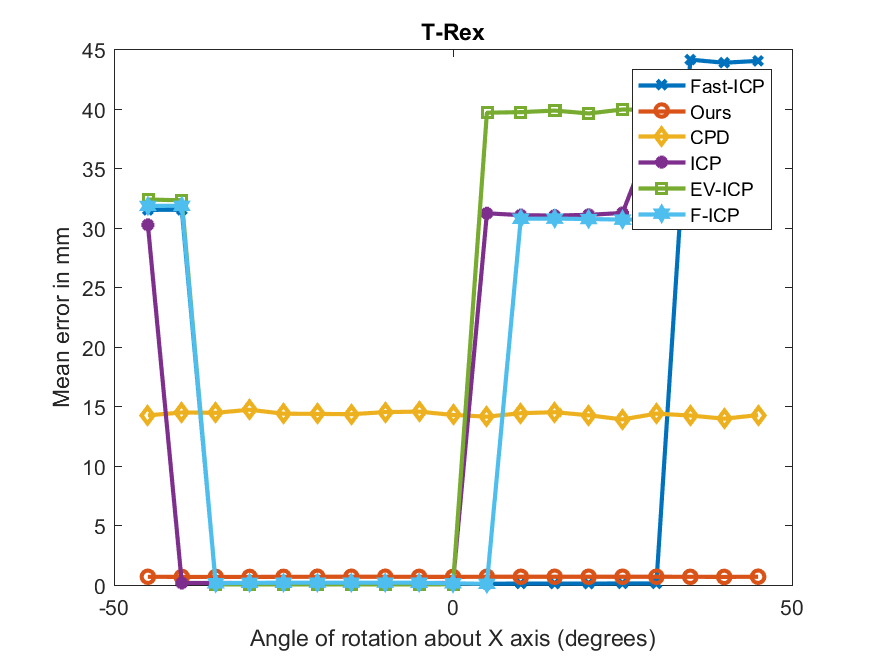}
\end{minipage}
\begin{minipage}[c]{0.33\textwidth}
\centering
\includegraphics[width=1\textwidth]{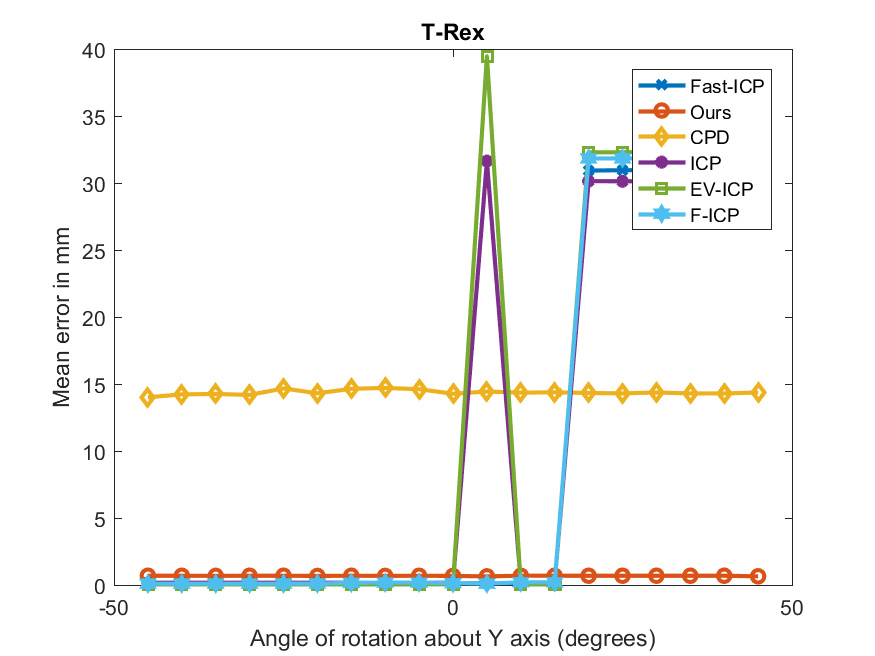}
\end{minipage}
\begin{minipage}[c]{0.33\textwidth}
\centering
\includegraphics[width=1\textwidth]{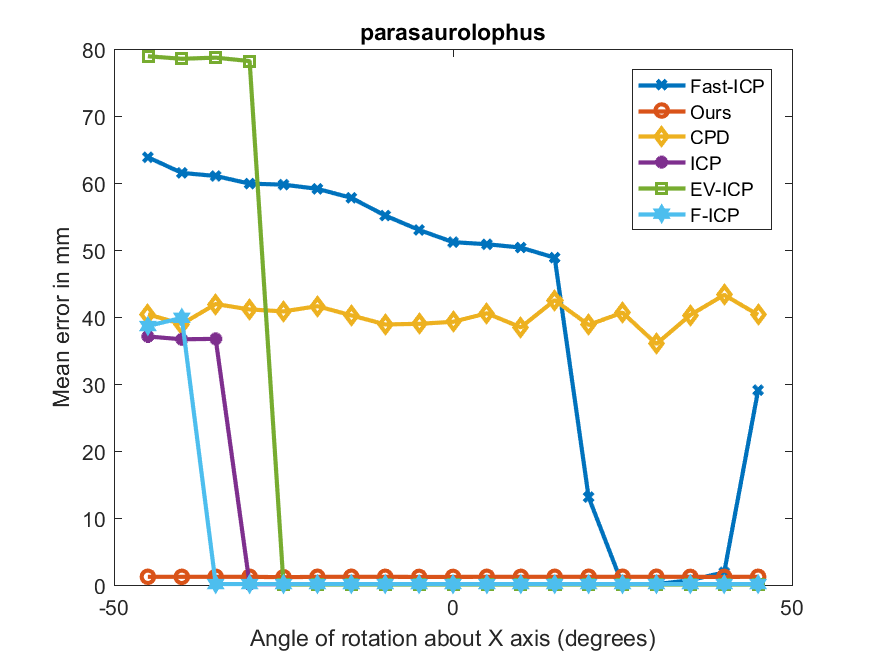}
\end{minipage}
\\
\begin{minipage}[c]{0.33\textwidth}
\centering
\includegraphics[width=1\textwidth]{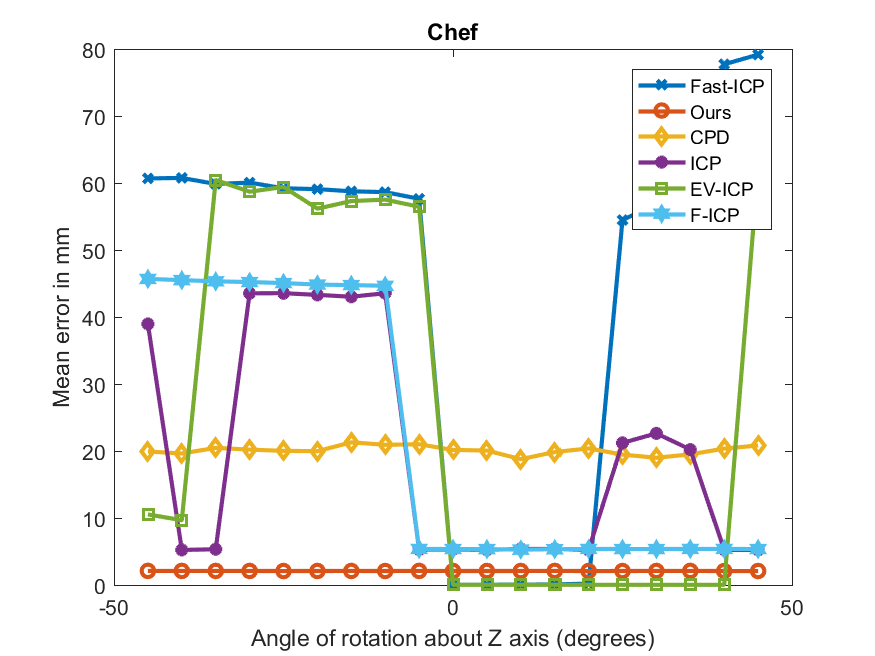}
\end{minipage}
\begin{minipage}[c]{0.33\textwidth}
\centering
\includegraphics[width=1\textwidth]{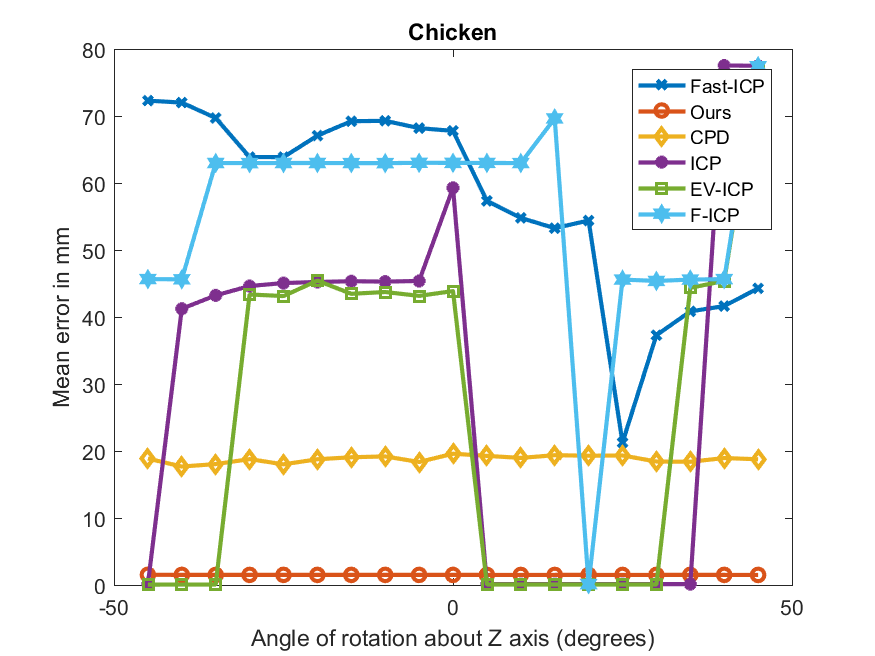}
\end{minipage}
\begin{minipage}[c]{0.33\textwidth}
\centering
\includegraphics[width=1\textwidth]{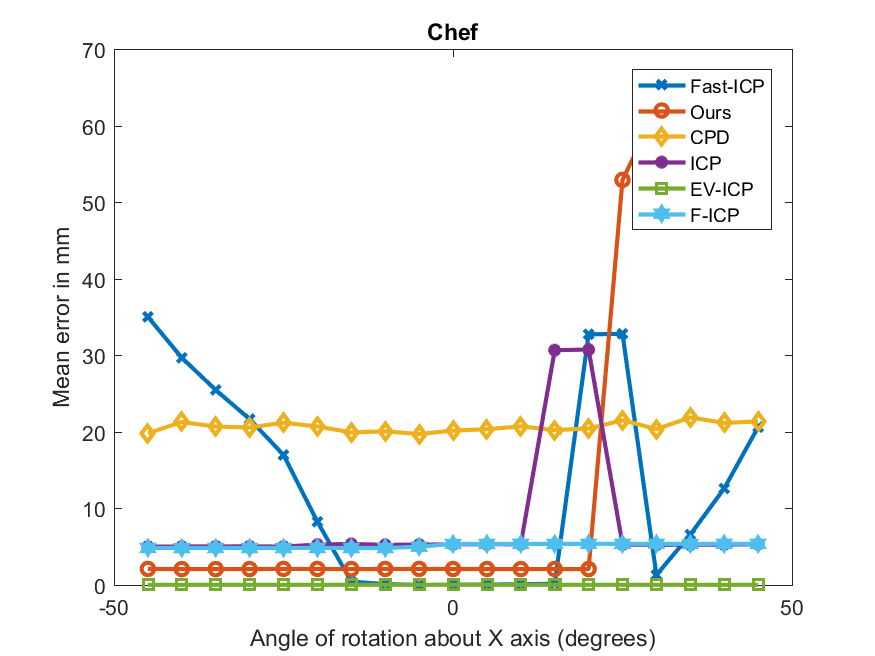}
\end{minipage}
\caption{Robustness to initialization: The ground truth alignment was used to synthetically add rotations and test different registration techniques. For each graph, the x-axis is the amount of rotation synthetically added, and the y-axis is the mean error in mm. }
\label{fig:registrationGraphs}
\end{figure*}

\begin{figure*}
\centering
\begin{minipage}[c]{0.33\textwidth}
\centering
\includegraphics[width=1\textwidth]{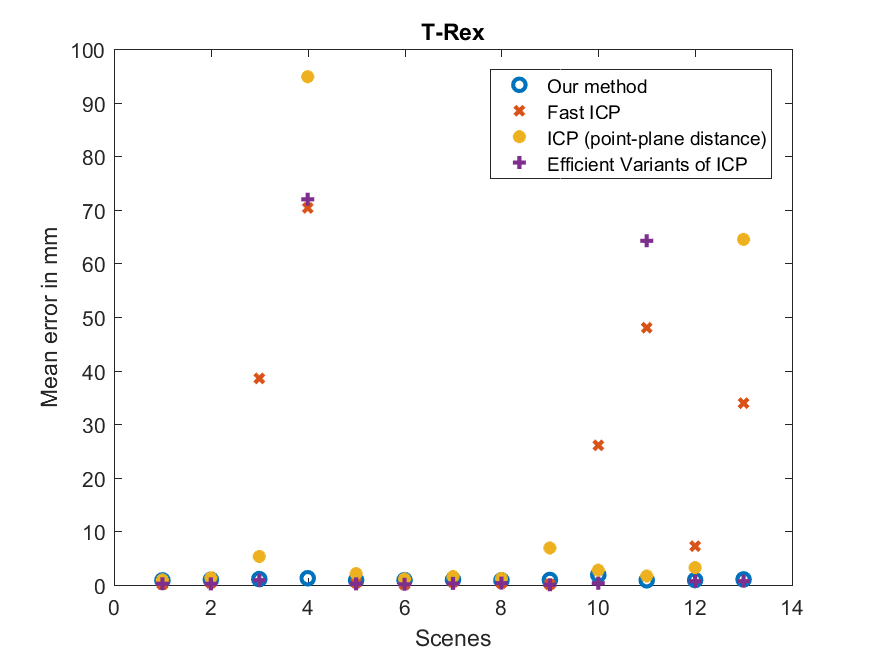}
\end{minipage}
\begin{minipage}[c]{0.33\textwidth}
\centering
\includegraphics[width=1\textwidth]{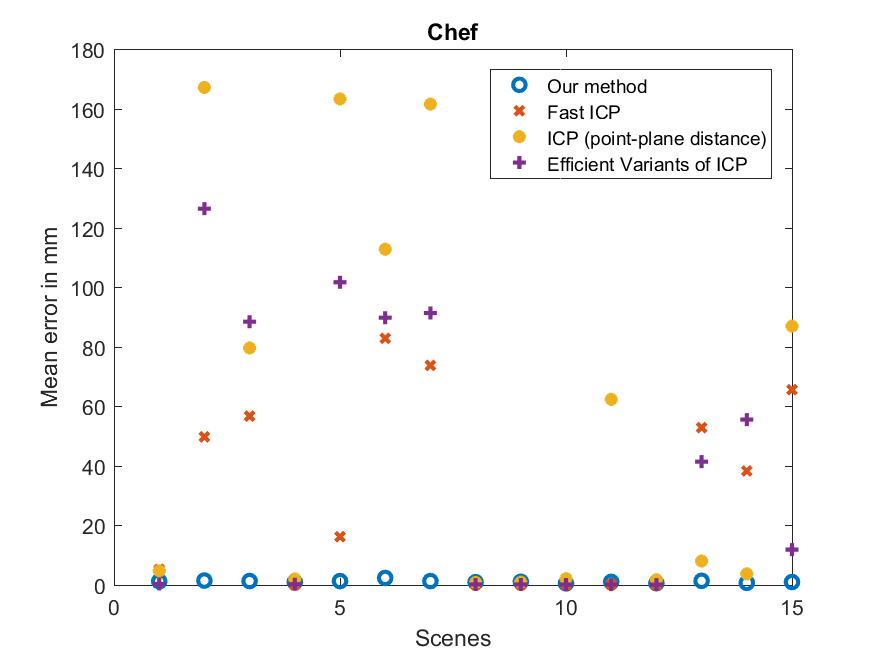}
\end{minipage}
\begin{minipage}[c]{0.33\textwidth}
\centering
\includegraphics[width=1\textwidth]{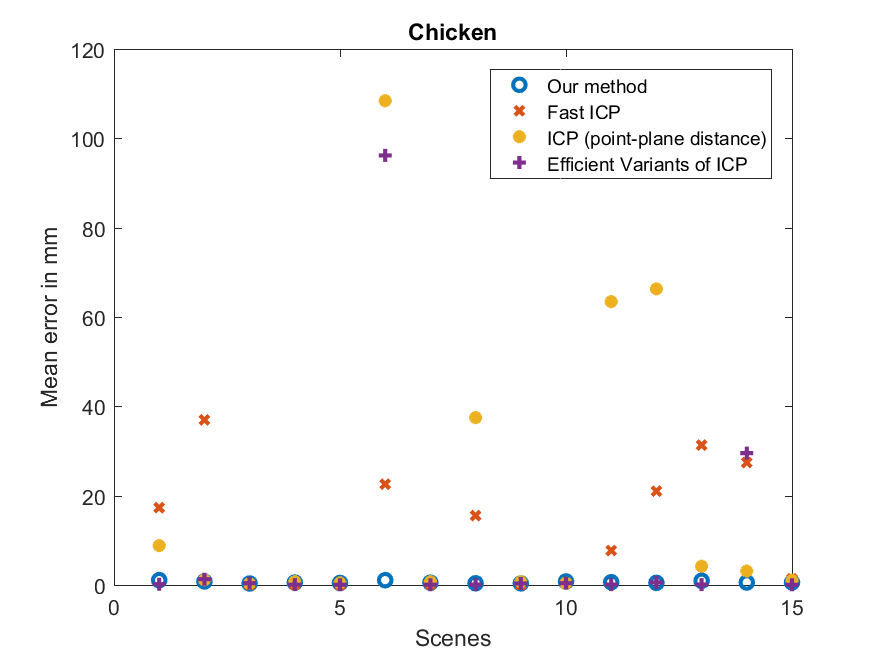}
\end{minipage}
\caption{Registration experiments on the first 15 scenes and 3 objects from the dataset provided by \cite{Mian2006,1677516}. The scenes were manually segmented to identify objects. For each object two different initialization were used and results from the one that had the least error after registration is shown here. The first initialization was aligning the centroids and the second initialization was aligning the centroids and a 180 degrees rotation of the model about the X-axis. Error was measured using the ground truth provided.}
\label{fig:comparisongraph}
\end{figure*}

Figure \ref{fig:registrationGraphs} shows a series of graphs for different models in the Range Images for 3D Modeling dataset, where we have introduced rotations about different axes in steps of $5^{\circ}$. In these graphs our algorithm (in orange) demonstrates stability over a wide range of rotations about different axes. Our algorithm consistently outperforms other techniques in terms of robustness to initialization. We have included a failure case where our algorithm finds a local minima in one of the models about one of the axes (bottom right corner of Figure \ref{fig:registrationGraphs}). This represents a case where the target shape is more radially symmetric than other objects in the scene. Additional results are shown in Figure \ref{fig:comparisongraph} and Table \ref{tab:toys}
\begin{table}
\centering
\caption{Table shows the success rate of the proposed approach when registration was run on all 50 scenes in the dataset \cite{Mian2006,1677516}. Two initialization were used as described in Figure \ref{fig:comparisongraph}. We say registration is successful if the mean error to the ground truth is $< 3mm$.}
\label{tab:toys}
\begin{tabular}{|c|c|}
\hline
  Model       & Success Rate  
  \\
  \hline
  T-Rex               & 0.80
  \\ 
  Parasaurolophus     & 0.89
  \\ 
  Chef                & 0.90
  \\ 
  Chicken             & 0.98
  \\ 
	\hline
\end{tabular}
\end{table}

\subsection{LiDAR  to CAD model Registration}
One of the main applications for registration techniques is aligning high quality CAD models to noisy sensor data for scene representation and robotics. Registering objects like valves and door handles is a key component of successfully navigating tasks like those in the DARPA Robotics Challenge.  This task is complicated by noise in sensor data. To illustrate the robustness of our method  we have aligned CAD models and LiDAR scans. The LiDAR scans come from the  Ford Campus Dataset \cite{Bao_CVPR2011_SSFM,Pandey:2011:FCV:2049736.2049742},  which uses a Velodyne 3D-LiDAR scanner. The CAD models were downloaded from the open source 3D model repository 3D Warehouse, and have been produced by different individuals, at different scales, resolutions, and level of detail. These models are not uniform, some contain internal mesh components like seats, some have additional body components like roof racks and larger wheels that may not reflect their real world counterparts, and these complicate the task of alignment.

To test our method we have manually identified vehicle models in the Ford Campus Dataset by looking through the photographs that are provided with the LiDAR scans. We have downloaded corresponding models from 3D warehouse to the best of our ability (within 2-3 model years for the vehicles we have selected). To test our registration technique we again segment the portion of the LiDAR scan which corresponds to the vehicle, and run our registration method. As this data contains no ground truth alignment parameters we report cloud-mesh distance using the CloudCompare utility \cite{cloudCompare}. The models we downloaded are at different scales (some in mm, some in feet and inches etc). Since the registration must handle scale, we have shown comparison only with CPD and fast CPD \cite{5432191}, as other ICP based methods do not handle scale well when the data is incomplete and sparse. We have also compared to CPD using our voxelization process described in \ref{voxel}. Results are shown in Table \ref{carTable}.

\begin{figure*}
\centering
\captionsetup{justification=centering}
\includegraphics[width=1\textwidth]{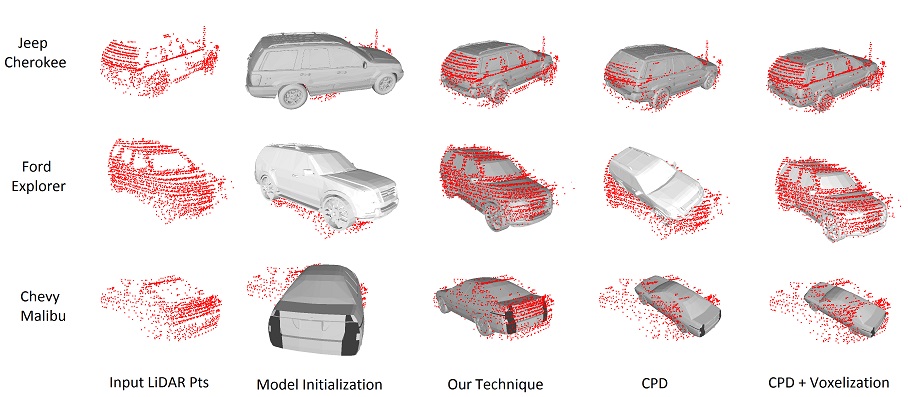}
\caption{LiDAR to CAD model alignment from the Ford Car Dataset using our method and CPD \cite{5432191}. Notice that with the more extreme initial misalignment of the Chevy Malibu, CPD fails to align while our method succeeds.}
\label{carFig}
\end{figure*}

\begin{figure*}
\centering
\includegraphics[width=1\textwidth]{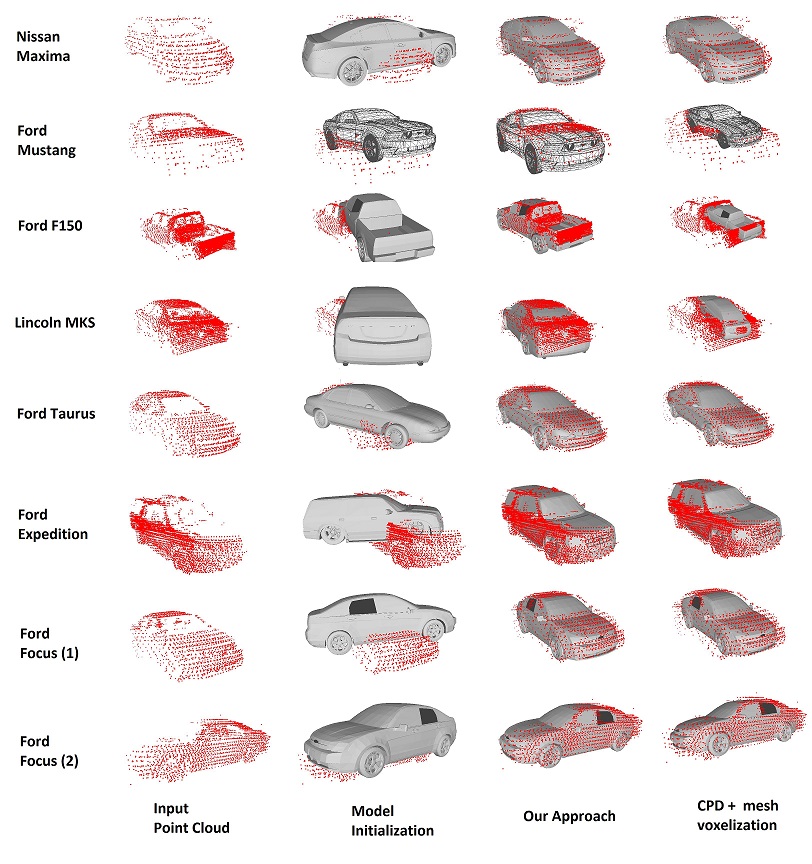}
\caption{Registration results on sparse, noisy, partial point sets of cars obtained from \cite{Bao_CVPR2011_SSFM,Pandey:2011:FCV:2049736.2049742}. }
\label{fig:cars}
\end{figure*}

\begin{table*}
\centering
\caption{Comparison of registration techniques on LiDAR to CAD models}
\label{carTable}
\begin{tabular}{|l|ll|ll|ll|ll|}
\hline
  Vehicle       & \multicolumn{2}{|l|}{Our Method}        & \multicolumn{2}{|l|}{CPD}      & \multicolumn{2}{l}{CPD (Fast)} & \multicolumn{2}{|l|}{CPD + Voxelization} \\ \hline
mesh     & mean dist         & max dist          & mean              & max      & mean           & max           & mean                   & max           \\
cherokee & 0.066379          & \textbf{0.634424} & \textbf{0.041224} & 0.706245 & 0.089925       & 0.813978      & 0.072922               & 0.754126      \\
explorer & 0.0587            & \textbf{0.335925} & 0.159617          & 1.01045  & 0.159631       & 0.998972      & \textbf{0.056963}      & 0.413368      \\
malibu   & \textbf{0.062977} & \textbf{0.30032}  & 0.204678          & 2.68191  & 0.21796        & 2.78548       & 0.229441               & 2.73506       \\
maxima   & 0.082867          & \textbf{0.461603} & 0.282231          & 2.25498  & 0.286168       & 2.2719        & \textbf{0.063284}      & 0.591919      \\
mustang  & \textbf{0.046983} & \textbf{0.349001} & 0.205525          & 1.78111  & 0.205479       & 1.79314       & 0.208201               & 1.96163       \\ 
Ford F150 & \textbf{0.0488093}          & \textbf{0.683794}         &-&-&-&-& 0.229246            & 2.97162      \\
Ford Focus 1 & \textbf{0.0598908}            & \textbf{0.490146}         &-&-&-&-& 0.069064            & 0.697694      \\
Ford Focus 2   & 0.0657001          & \textbf{0.526913}          &-&-&-&-& \textbf{0.0649599}            & 0.536582      \\
Ford Taurus   & \textbf{0.0375229}          & \textbf{0.461603}         &-&-&-&-& 0.0423505            & 0.582141      \\
Ford Expedition   & 0.061217          & 0.463239         &-&-&-&-& \textbf{0.0589582}            & \textbf{0.403849}      \\
Lincoln MKS  & \textbf{0.0589106}          & \textbf{0.574546}         &-&-&-&-& 0.174584            & 2.59754 \\ 
\hline         
\end{tabular}
\end{table*}

These results show that our method outperforms CPD in most circumstances. We consistently achieve lower mean and max distance from the cloud to mesh after applying our alignment. Figure \ref{carFig} and \ref{fig:cars} show the input point cloud, our alignment and the alignment achieved by CPD. The mean cloud to mesh distance from our method is approximately the same as the 5cm error range of the LiDAR used to capture the data. 

\subsection{LiDAR to Camera Alignment}
We have extended the formulation of our registration technique to work on not only 3D to 3D alignment, but projective 2D to 3D alignment by ray-point registration. This extension allows for automatic and semi-automatic LiDAR to camera calibration, and has applications in sensor fusion, and 3D reconstruction. We demonstrate this technique by aligning and re-projecting LiDAR points onto both color and and thermal images. We have developed this technique for validating stereo reconstruction in two modalities using LiDAR to generate ground truth.

We have used a Trimble GX Advanced TLS LiDAR, which captures 5000 3D points per second at $<2mm$ error at 50M. The LiDAR is mounted adjacent to a color and thermal camera system. The color camera is a Point Grey Flea2 capturing at 1280 x 960. The thermal camera is a long wave infrared Xenics Gobi 640-GigE, which captures at 640 x 480. The cameras have similar fields of view, and we scan an area slightly larger than what either camera sees. The cameras are calibrated using the method outlined in  \cite{7351702} using a heated ceramic backed calibration pattern, allowing for both cameras to be calibrated simultaneously. 

Calibration of the LiDAR to camera is a process of identifying the transformation from the camera center to the coordinate center of the LiDAR. We treat this as a process of registering 3D rays to 3D points, and solve for the alignment using the technique outlined in Section \ref{sec:2d_3d}. To do this we generate a set of rough correspondences in the form of automatically generated points via edges or roughly drawn regions for semi automatic registration.

\begin{figure}[h]
\captionsetup{justification=centering}
\includegraphics[width=0.49\textwidth]{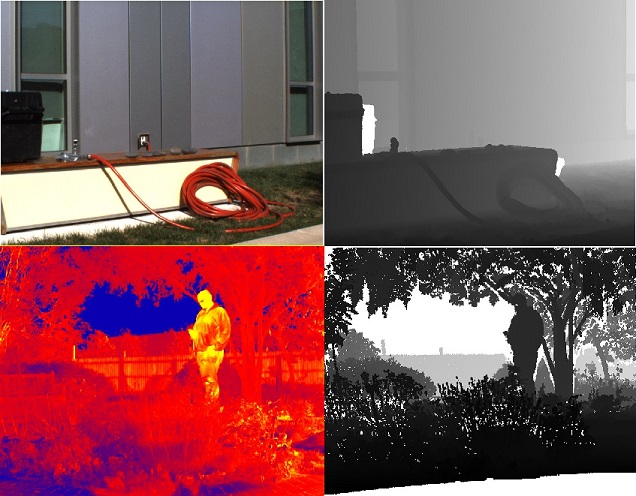}
\caption{Sample color and thermal images and generated depth maps from calibration using our technique}
\label{cameraLidarFig}
\end{figure}

To automatically generate rough correspondences we use image edges and depth discontinuities. We project the 3D LiDAR points to a 2D depth image using the camera parameters. We then compute an edge mask of both the depth image and a bilateral filtered camera image using Canny edge detection \cite{Canny:1986:CAE:11274.11275}. Furthermore, this process works exceptionally well in thermal images, because texture has little effect on the image intensity, and most of the edges correspond to object boundaries and material changes. These edges are used to generate a set of 3D points and rays for alignment. 2D points in the camera edge-map are used to create a set of rays originating at the camera center and intersecting the image plane through the edge map. The depth edge map is dilated, and all the 3D points that project to the new edge-map are obtained. These rays and 3D points are then used for registration. The resulting transformation allows us to project 3D points from the LiDAR onto the camera, and generate new depth images from the camera's perspective as shown in Figure \ref{cameraLidarFig}.

\begin{figure}[h]
\centering
\includegraphics[width = 0.49\textwidth]{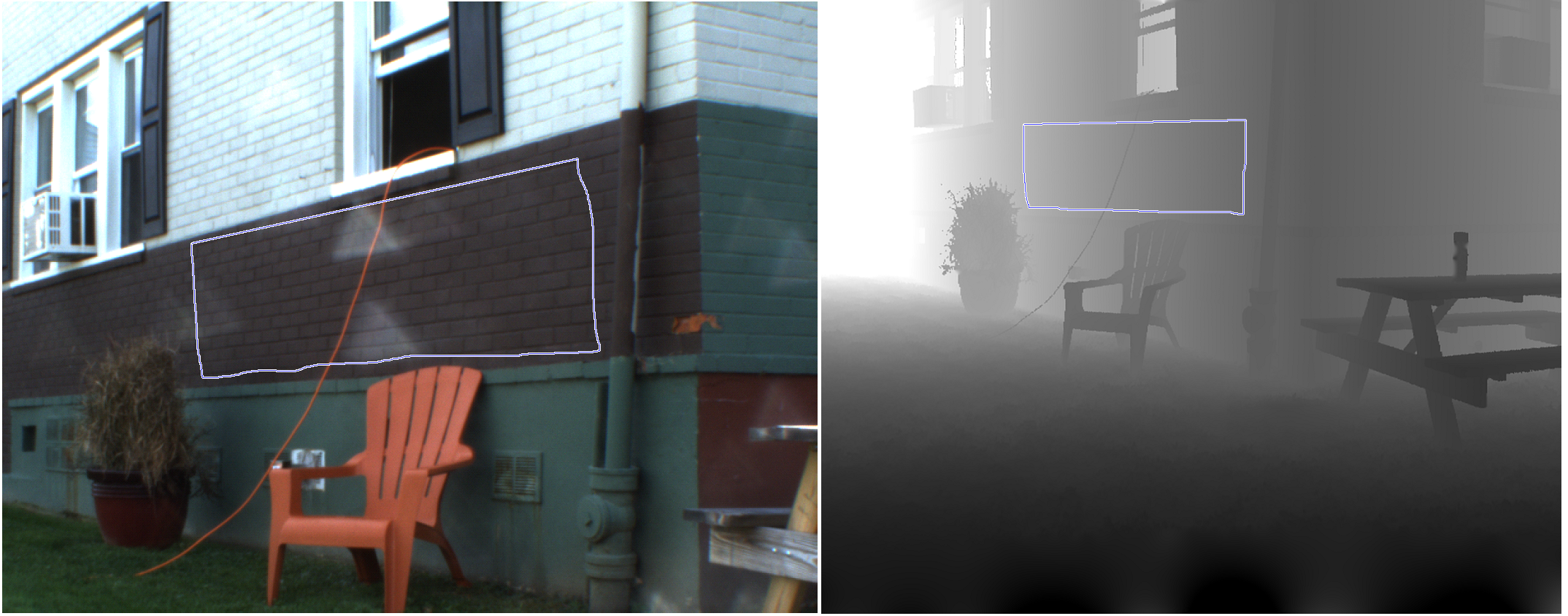}
\caption{Our GUI allows user to highlight areas in the camera image and depth image to manually create rough correspondences}
\label{fig:GUI}
\end{figure}

In scenes with dense and contrasting texture the camera edge image can be exceedingly noisy, and this can lead to a large number of potential local minimums. To combat this we have extended the above method to include manual rough correspondences. We have developed a small GUI which allows users to draw regions on both the camera image and the depth image to create rough correspondences. It is not mandatory for the hand drawn components to be exact same area or even to have the same number of components in each image. Figure \ref{fig:GUI} shows the GUI which allows users to highlight regions in the camera image and depth image to generate rough correspondences. 

\begin{figure}[h]
\centering
\includegraphics[width = 0.4\textwidth]{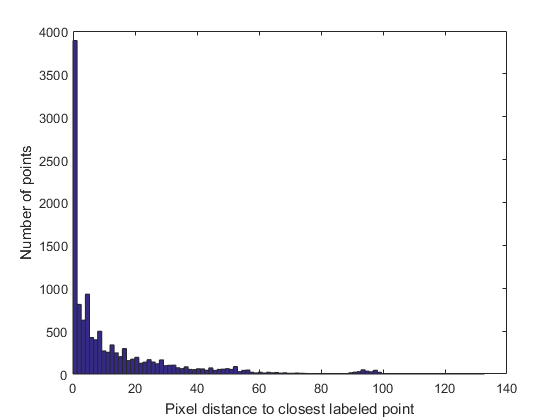}
\caption{Histogram of errors across 5 different ray-point alignment tasks with different transformations}
\label{errorHistogram}
\end{figure}

To validate this approach we measure pixel distance from reprojected 3D points to their closest labeled point in the 2D image. We report the errors over 5 different alignments in the form of a histogram in Figure \ref{errorHistogram}. The large majority of points report error of under 3 pixels. The outliers with large errors are caused by points that have no correspondence and (correctly) lie outside the image when projected.

\begin{figure*}[h]
\begin{minipage}[c]{0.275\textwidth}
\centering
\includegraphics[width=1\textwidth]{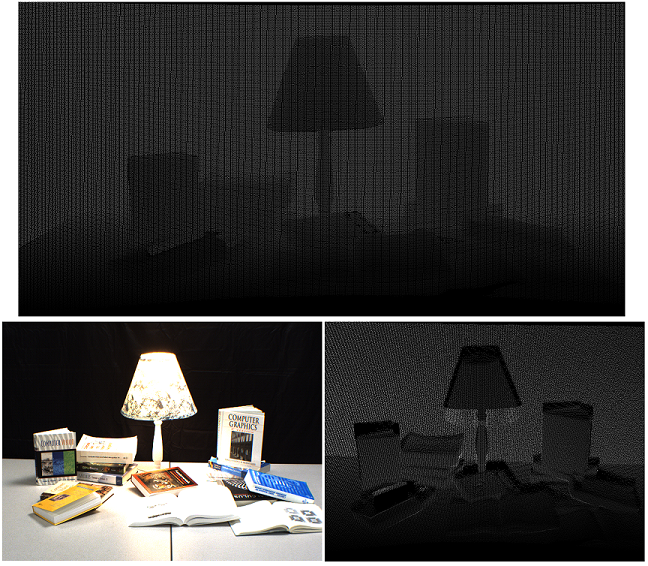}
\\
\ 
\end{minipage}
\hfill
\begin{minipage}[c]{0.4\textwidth}
\centering
\includegraphics[width=1\textwidth]{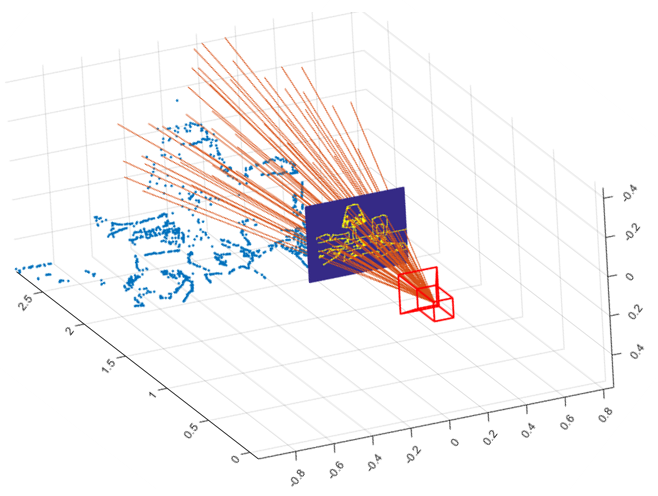}\\

(i)
\end{minipage}
\begin{minipage}[c]{0.3\textwidth}
\centering
\includegraphics[width=0.8\textwidth]{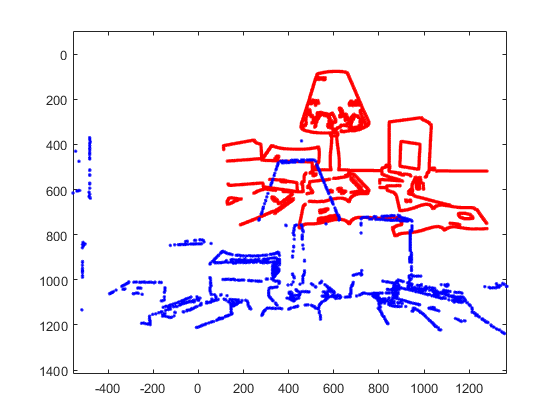}\\
\includegraphics[width=0.8\textwidth]{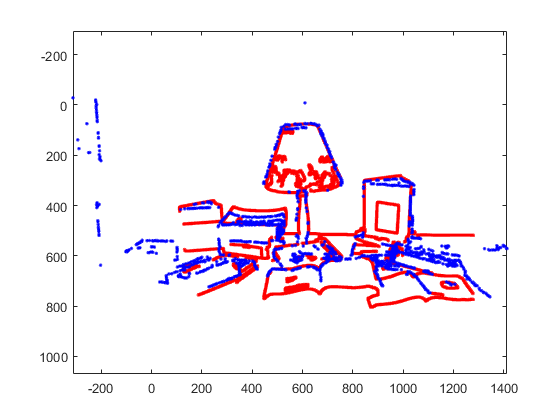}\\
(ii)
\end{minipage}
\\
\begin{minipage}[c]{0.275\textwidth}
\centering
(A)
\end{minipage}
\hfill
\begin{minipage}[c]{0.7\textwidth}
\centering
(B)
\end{minipage}
\caption{Lidar to Camera alignment: (A) Results of alignment. Top row: Input depth map. Bottom row left: Input color image. Bottom row right: Depth map aligned to the camera image. (B)The alignment process. (i) Plot showing the 3D points and rays used for alignment. (ii) Top row: Initial alignment. Bottom row: Final alignment result using our approach}
\label{fig:lidar2camalignment}
\end{figure*}
\section{Conclusion}
\label{sec:con}
In this work we have presented a robust technique for registration that does not require correspondences, effectively handles scale, and is invariant to many transformations that cause traditional approaches to fail. The proposed approach based on fuzzy correspondences maximizes the overlap and proximity of points between the target and source shape. We have demonstrated this technique on data from public datasets and compared it against widely used approaches. Our technique outperforms the other methods in registration tasks. With noisy sensor data we have shown that our technique can align models at different scales to within a close range of the sensor's error margins. We have also extended the technique to handle the problem of ray-point registration for the application of LiDAR to camera alignment, and we have developed fully automatic and semi-automatic approaches to calibration that do not require a calibration object, and work in different image modalities.

{\small
\bibliographystyle{ieee}
\bibliography{egbib}
}

\end{document}